\documentclass[10pt,twocolumn,letterpaper]{article}

\usepackage{cvpr}              % To produce the CAMERA-READY version
\usepackage{multirow}
\usepackage{xr}
\usepackage{amsmath}
\usepackage{amssymb}
\definecolor{cvprblue}{rgb}{0.21,0.49,0.74}
\usepackage[pagebackref,breaklinks,colorlinks,allcolors=cvprblue]{hyperref}

\def\paperID{38}% *** Enter the Paper ID here
\def\confName{CVPR}
\def\confYear{2025}

\title{Distilling Normalizing Flows}

\author{
    Steven Walton\textsuperscript{1,2,4}\footnote{swalton2@uoregon.edu}\quad
    Valeriy Klyukin\textsuperscript{3}\qquad
    Maksim Artemev\textsuperscript{4}\\
    \hspace{-1em}Denis Derkach\textsuperscript{3}\quad\;\;
    Nikita Orlov\textsuperscript{4}\qquad
    Humphrey Shi\textsuperscript{2,4}
    \\
    {\textsuperscript{1}\small University of Oregon}\qquad
    {\textsuperscript{2}\small SHI Lab @ GaTech}\qquad
    {\textsuperscript{3}\small HSE University}\qquad
    {\textsuperscript{4}\small Picsart AI Research}
}

\begin{document}
\maketitle
\begin{abstract}
    Explicit density learners are becoming an increasingly popular technique for generative models because of their ability to better model probability distributions. 
    They have advantages over Generative Adversarial Networks due to their ability to perform density estimation and having exact latent-variable inference.
    This has many advantages, including: being able to simply interpolate, calculate sample likelihood, and analyze the probability distribution.
    The downside of these models is that they are often more difficult to train and have lower sampling quality.
    
    Normalizing flows are explicit density models, that use composable bijective functions to turn an intractable probability function into a tractable one. 
    In this work, we present novel knowledge distillation techniques to increase sampling quality and density estimation of smaller student normalizing flows.
    We seek to study the capacity of knowledge distillation in Compositional
    Normalizing Flows to understand the benefits and weaknesses provided by
    these architectures.
    Normalizing flows have unique properties that allow for a non-traditional forms of knowledge transfer, where we can transfer that knowledge within intermediate layers. 
    We find that through this distillation, we can make students significantly smaller while making substantial performance gains over a non-distilled student.
    With smaller models there is a proportionally increased throughput as this is dependent upon the number of bijectors, and thus parameters, in the network.
% \keywords{Generative Modeling, Normalizing Flows, Knowledge Distillation, Knowledge Transfer}
\end{abstract}
    
\section{Introduction}
\label{sec:intro}

Generative models are a popular form of deep learning, due to their ability to generate convincing samples from a probability distribution.
The de facto choice for image sampling are Generative Adversarial Networks (GANs)~\cite{goodfellow2014generative}.
While GANs have been shown to produce high-quality samples that look similar to the samples from target distribution, they do not faithfully reproduce the target probability distribution~\cite{Nagarajan2019TheoreticalII}.
Arora~\etal~\cite{arora2018do} demonstrate this by looking at the Birthday Paradox and analyzing the diversity of images.
Anyone who has used popular ``this $x$ does not exist'' websites will quickly notice how similar some images are to those they were trained on.
%This is mostly due to the limited latent representation that GANs have.
This is due to that GAN's non-exact latent-variable inference, requiring them to have a smaller latent representation and thus being implicit density learners.
On the other hand, Normalizing Flows have full latent representations, not using any encoders or decoders as are used in GANs or VAEs, and allow them to learn tractable density functions. 
Not only may researchers and practitioners want to produce high quality samples, but some may also want to perform density estimation and analyze the target distribution. 
Explicit density learners, such as Normalizing Flows, are able to accomplish all of these tasks.
The major challenge is that it is substantially more challenging to learn all this extra information and require larger latent representations.
Variational autoencoders~\cite{rezende2015variational} try to solve this by attempting to learn only a subset of the latent space, reducing the dimensionality of the problem to a set of approximated principle components.
On the other hand, there are autoregressive models (AR)~\cite{wavenet,germain2015made}, normalizing flows (NFs)~\cite{rezende2015variational}, and deep energy-based models (EBMs)~\cite{lecun2005loss,sohldickstein2015deep}, which attempt to learn all of the high dimensional data and likelihood information.
While these models reproduce the target distribution more faithfully, they are much more difficult to train and require significantly more computational resources. 

The advantage of NFs over other explicit density models is that they are simple in structure and fully tractable throughout the entire model.
They accomplish their learning by composing bijective functions (equations whose maps are one-to-one and onto).
This means that NFs perform a change of variables operation, changing an intractable probability distribution into a tractable one, typically Gaussian.
NFs do this through the use of a learned change of variables mapping.
The advantage of this is that once we have found the proper simpler density function, we can easily analyze it and gain all the statistical advantages of a tractable probability density function.

Normalizing Flows are particularly difficult to train due to their invertible nature, requiring \textit{diffeomorphic} compositions and thus we require a tractable Jacobian determinants to preform inference through the backwards pass.
While tractable Jacobian determinants are not required for forward inference, it is the bidirectional nature that makes NFs particularly unique. 
This aspect has been the major challenge of NFs, as it is difficult to find bijective functions that are both efficient and highly expressive~\cite{papamakarios2021normalizing,kobyzev2021}. 
Current techniques lead to a high number of parameters, meaning that they also tend to be inefficient during inference. 
This work focuses on using knowledge distillation~\cite{hinton2015distilling} to train smaller models through a student-teacher paradigm.
While smaller models are more difficult to train to high accuracy, they perform faster inference due to the reduction in the number of parameters.
NFs have unique properties that allow for unique types of knowledge distillation techniques compared to other more traditional models.
Specifically, since NFs focus on a change of variables, they learn full latent representations through intermediate layers accurately, which can be distilled between models of similar architectures. 

The study of distillation within Normalizing Flow architectures is understudied.
In this paper, we explore this unique property and demonstrate its improvements 
in student models.
In this manner we hope to provide a general framework in which distillation may
be applied to these networks.
We do not seek to find optimal settings or methods of distillation, but to study
the capacity of these models to distill knowledge.
More specifically, to do so by taking advantage of the unique properties of
these architectures and not found within other architectures, such as GANs or
VAEs.
Furthermore, we believe the insights gained from this work may inspire research
for architectures such as diffusion
models~\cite{NEURIPS2020_4c5bcfec,song2021denoising} which share some, but not
all, properties.

Our main contributions in this paper are:
\begin{itemize}
    \item We propose novel knowledge distillation techniques for normalizing flows, demonstrating their effectiveness on smaller flow models.
    \item We demonstrate the effects of different distillation techniques on these normalizing flows. 
    \item We demonstrate that these methods can be used on various datasets, including tabular datasets and common image datasets.
    \item We provide a foundation for studying distillation methods applied to
        Normalizing Flows.
\end{itemize}

\section{Related Work}
\label{sec:related}

\subsection{Normalizing Flows}
\label{sec:nf}

Normalizing Flows are a type of explicit density model, where instead of attempting to just generate samples similar to our target distribution, the objective is to model the entire distribution.
This can be accomplished through a change of variables method, where a transformation $f$ is used that is both invertible, and where $f$ and $f^{-1}$ are both differentiable, meaning that $f$ is \textit{diffeomorphic}.
This change of variables can be expressed as 
\begin{equation}
    \label{eq:diffeo}
    p_x(\boldsymbol{x}) = p_u(\boldsymbol{u}) \left| \text{det}J_f(\boldsymbol{u}) \right|^{-1} \vspace{2em} 
\end{equation}
where $\boldsymbol{u}=f^{-1}(\boldsymbol{x})=g(\boldsymbol{x})$. $p_x$ represents the distribution being modeled, and $p_u$ represents the learned distribution.
In other words, this change of variables is expressed as a learned distribution
times the absolute value of the inverse of the Jacobin determinant, $\text{det}J$.
This inverse Jacobian determinant is why Normalizing Flows are computationally expensive.

Because these functions are diffeomorphic, they are also composable. 
Therefore, with an arbitrary transformation $f_i$, a normalizing flow can be created, which is expressed as
\begin{equation}
\label{eq:composition}
    \boldsymbol{f} = \boldsymbol{f}_1 \circ \boldsymbol{f}_2 \circ \dots \circ \boldsymbol{f}_k,
\end{equation}

We refer to the number of composable functions as the depth of the flow network, $k$.
The Jacobian determinant of the entire network can be calculated by the product
of the Jacobian determinant at each layer.
\begin{equation}
\label{eq:composition_det}
    \det J_{\boldsymbol{f}}(\boldsymbol{x}) = \prod_{i=1}^n \det J_{\boldsymbol{f}_i}(\boldsymbol{x}_i)
\end{equation}
Triangular decomposition is frequently used to solve the Jacobian determinant, as the determinant of a triangular matrix is trivial to compute.

With this, any sample $x$, from the intractable distribution, $p_x$, can be fed into the model (function $f$), and transformed to a tractable distribution $p_u$. 
Then, from the tractable distribution, $p_u$, new samples can be generated by inverting the flow. %, $p_x(\boldsymbol{x}) \approx g(\boldsymbol{u})$.
We refer to the inverse, $f^{-1}$, as $g$ to denote that this direction is a generator, where one can sample from a distribution similar to $p_x$. 

There are several types of normalizing flow architectures: Planar and Radial Flows introduced by Rezende and Mohamed \cite{rezende2015variational};
 Coupling Flows, proposed by Dinh et al. \cite{dinh2016density};
 Autoregressive Flows, presented by Kingma et al. \cite{kingma2016improving};
 Continuous Flows, suggested by Chen et al. \cite{chen2018neural}.

Each of these has different trade-offs between expressiveness and simplicity to calculate, with Planar and Radial Flows being the most computationally complex.

Since Coupling and Autoregressive Flows are the most common, we focus on the most popular representatives of each: GLOW (Kingma and Dhariwal \cite{kingma2018glow}) and Masked Autoregressive Flow (Papamakarios et al. \cite{papamakarios2017masked}).

Normalizing flows are also included in the class of likelihood-based learners, or neural networks that attempt to solve the maximum likelihood of the distribution.
Because of this, one can use any metric to minimize the discrepancy between the target distribution and the learned one.
Typically the Kullback-Leibler (KL) divergence is used, which uses entropy between the two distributions to determine the difference.
Therefore, the loss for a sample $\boldsymbol{x}$ and parameter $\boldsymbol{\theta}$ can be expressed as:
\begin{align}
    \mathcal{L}(\theta) &= D_{KL}\left[p_x(\boldsymbol{x}) || p_u(\boldsymbol{x};\boldsymbol{\theta})\right]\\
        &= \sum p_x(x)\log\left(\frac{p_x(x)}{p_u(x)}\right)
\end{align}
While this might not always be possible, there are many alternative equivalents such as the reverse KL-Divergence and Evidence Lower Bound (ELBO), which can be used in cases where sampling from the target distribution may not be possible.

Additionally, when training on discrete variables, such as images or text, the dataset typically needs to be dequantized to transform them into continuous distributions.
While there are methods to work on discretized data~\cite{hoogeboom2019integer,lindt2021discrete}, like images, the common practice is to first dequantize the data.
%That is, discrete variables need to be transformed into continuous one, it must be ensured that the volume of the probability density is 1 ($\int p(\boldsymbol{x})dx =1$).
%This ensures that a probability density is actually being modeled, which is essential to the learning process of Normalizing Flows.
For images, the image values can be converted from integers in $[0, 255]$ to real numbers in $[0, 1]$, noise is added to continuize (dequantize) the data and transform it into a tractable density, typically Gaussian.
This also means that the dequantization process is, approximately, invertible, which makes this equivalent to another flow step.

\subsubsection{GLOW}
\label{sec:glow}

GLOW is an extension of the RealNVP~\cite{dinh2016density} architecture, which uses affine coupling layers.
Among the innovations proposed in the original paper were the activation normalization layer (ActNorm) and the $1\times1$ invertible convolution layer.
ActNorm has the following form:
\begin{equation}
\label{eq:actnorm}
    \forall i, j: \boldsymbol{x}_{i, j} = \boldsymbol{s} \odot \boldsymbol{z}_{i, j} + \boldsymbol{b}
\end{equation}

$\boldsymbol{b}$ and $\boldsymbol{s}$ are initialized so that the mean and standard deviation, respectively, of the first mini-batch of data is $0$ and $1$.
An invertible $1 \times 1$ convolution is arranged as follows:
\begin{equation}
\label{eq:1x1conv}
    \forall i, j: \boldsymbol{x}_{i, j} = W \boldsymbol{z}_{i, j}
\end{equation}
where $W \in \mathbb{R}^{c \times c}$ is initialized as a square orthonormal matrix of the same size as the number of channels, $c$.
%In this case, the matrix $W$ is parametrized through LU-decomposition:
With this setup we can more easily calculate the Jacobian determinant through
LU-decomposition.
We can further decompose U by giving it a unitary diagonal and adding it to a
diagonal matrix where entries are that of the original U matrix:
\begin{equation}
\label{eq:1x1conv_matrix}
    W = PL(U + diag(\boldsymbol{s})),
\end{equation}
where P is a permutation matrix, L is a lower triangular matrix with unit diagonal, U is an upper triangular matrix.% with zero diagonal and $\boldsymbol{s} \in \mathbb{R}^c$.

%This makes it trivial to calculate the Jacobian determinant as we just need to take the trace of U.
%We can further decompose U into a unit diagonal and add $\text{diag}(\boldsymbol{s})$, giving a simplified log Jacobian Determinant equation: %equation~\ref{eq:1x1conv_matrix}.
By pulling the trace of U out, this allows for a simplified calculation of the
log Jacobian determinant, since the determinant of a diagonal matrix is the sum
of its trace.
\begin{equation}
\label{eq:ldjconv}
    \log(\text{det}(J)) = h \cdot w \cdot \sum\log|s|
\end{equation}
where $h$ and $w$ are the height and width (number of pixels) of the input image. 
We note that while LU-Decomposition significantly reduces the computational complexity, that this function is still rather costly to calculate.

Coupling layers may be expressed in several forms, often additive or affine. 
These layers are highly expressive flows that operate on a disjoint partition.

\subsubsection{Masked Autoregressive Flow}
\label{sec:maf}
%In Masked Autoregressive Flow 
(MAF), models each element of the vector $\boldsymbol{x}$ as a conditioned function $h(\cdot, \theta)$:
\begin{equation}
\label{eq:maf}
    \boldsymbol{x}_i = h(\boldsymbol{z}_i, \Theta_i(\boldsymbol{z}_{1:i-1}))
\end{equation}
The determinant of the Jacobian for such a transformation is a lower triangular matrix since $\boldsymbol{x}_i$ depends only on $\boldsymbol{z}_{1:i}$, thus making it trivial to calculate.
The calculation of $\boldsymbol{x}$ can be performed with a single network pass by masking the $\Theta$ and $h$ network layers.

\subsection{Knowledge Transfer and Distillation}
With knowledge transfer one can incorporate additional loss information into a model being trained by using output from an already pretrained model.
Typically we do this from a model that is larger, called a teacher, that will \textit{distill} the information to a smaller model, called a student. 
This can help the student model both learn faster and avoid local minima that it might typically get trapped in. 
There are several schemes of KD:
\begin{itemize}
    \item Response-based KD: constraining student to produce outputs similar to that of the teacher's.
    This method is the standard for KD of discriminative models and was first proposed by Hinton et al. \cite{hinton2015distilling}.
    \item Feature-based KD: constraining student to have internal representations similar to the teacher's. Similar method was proposed by Romero et al. \cite{romero2015fitnets} for general knowledge distillation framework.
    \item Relation-based KD: training a student to replicate teacher's relationships between different layers or different data points.
\end{itemize}

A recent work by Baranchuk~\etal~\cite{baranchuk2021distilling} proposed using knowledge distillation from a conditional normalizing flow to teach a feed-forward based flow like SRFlow and WaveGLOW. 
This work used response-based knowledge distillation, similar to that presented by Hinton et al. 
This demonstrates that conditional flows can be trained with the standard knowledge distillation techniques, only using the information from the outputs of the teacher and student, but does not demonstrate that such distillation techniques can be used on unconditional flows, invertible flows, nor can feature-based or relation-based knowledge distillation techniques be used.

\label{sec:method}
\begin{figure*}
    \centering
    \includegraphics[width=0.75\textwidth]{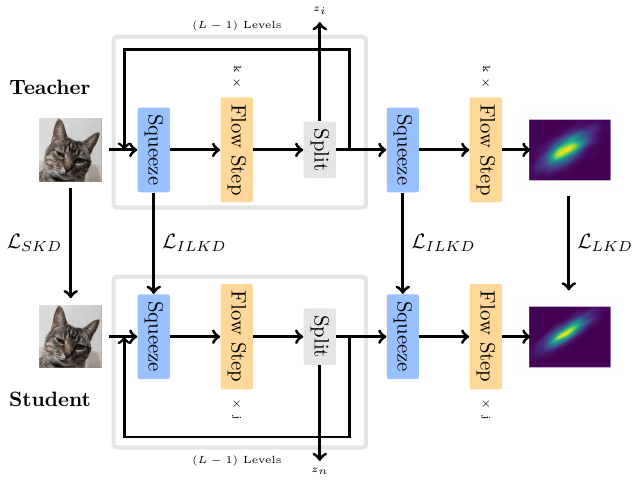}
    \caption{Illustration of different distilling mechanisms using a glow like multi-level architecture. $\mathcal{L}_{SKD}$ shows the synthesized knowledge distillation, $\mathcal{L}_{ILKD}$ shows the intermediate latent knowledge distillation, used at every squeeze step (same as ours) and $\mathcal{L}_{LKD}$ shows the latent knowledge distillation. Here our teacher has $k$ steps (depth) of flow and our student has $j$, where $j<k$. Note that they both have the same number of levels.}
    \label{fig:distill}
\end{figure*}
\subsection{Probability Density Distillation}
\label{sec:pdd}

Probability Density Distillation, PDD, was first introduced by van den
Oord~\etal~\cite{oord2018parallel} as a potential solution to WaveNet's~\cite{oord2016wavenet} poor sampling performance, thus it became the de facto standard in the waveform modulation due to the significant increase in sampling rates. 

The original WaveNet (WN) model is autoregressive, which benefits from the PDD training process, as the model can efficiently run in parallel due to its convolutional nature.
However, while generating samples, input features for each layer must be first drawn from the previous feature layer, making parallel inference impossible.

% fix???
The PDD method aims to translate learned distribution from an easy-to-train autoregressive model to the easy-to-parallelize Inverse Autoregressive Flow (IAF)~\cite{kingma2016improving}.
Since both of the described models can do density estimation, the proposed way to distillate is to constrain the IAF's output to match the pre-trained WaveFlow model:

\begin{equation}
\label{eq:ppd}
    D_\text{KL}(P_\text{IAF} \| P_\text{WN}) = H(P_\text{IAF} \| P_\text{WN}) + H(P_\text{IAF})
\end{equation}

where $H(P_\text{IAF} \| P_\text{WN})$ and $H(P_\text{IAF})$ denotes cross-entropy and entropy respectively.

% написать подробнее про лоссы и как учится студент
% обьяснить почему мы не используем power/perceptual loss из статьи

The PDD method quickly became a baseline solution for training small flows for waveforms modulation and inspired a variety~\cite{ping2018clarinet,prenger2019waveglow} of other works. Recent work presented by Hoogeboom et al.~\cite{hoogeboom2021autoregressive} takes a different approach by incorporating techniques from diffusion models to also increase efficiency. 

% додумать как вставить сюда реф к >On Difficulties of Probability Distillation

\section{Method}

Normalizing flows are unique in their structure due to the composition of composable diffeomorphic functions.
This means that we can transform information in many unique ways as compared to other types of deep neural networks. 
Because normalizing flows operate by a change of variables transformation, we can assume certain properties at different depths of the network.
Namely, if we use different levels, in a hierarchical structure, like glow, we may want the student and teacher to share similar latent distributions at those points. 
Similarly, since we are able to model probability density functions, we know that we want the student and teacher to share this same information.
We discuss these techniques in the following subsection.
Additionally, we note that these techniques are not limited to feed-forward students nor do they require conditional inference.
Thus these techniques extend the work of Baranchuk~\etal~\cite{baranchuk2021distilling}

%\subsection{knowledge distillation}

\subsection{Latent Knowledge Distillation}

A basic version of knowledge transfer for normalizing flows is to teach the student's latent distribution to be similar to the teacher's.
This is known as \textit{Latent Knowledge Distillation} (LKD).
%Essentially we can accomplish this by using a $l_1$ or $l_2$ loss between the
Loss can be calculated between
two latent vector representations, as shown in equation~\ref{eq:LKD}.
This loss may be arbitrarily, but we use a $L_1$ loss for our experiments.
\begin{equation} \label{eq:LKD}
    \mathcal{L}_\text{LKD}(t, s, x) = \mathcal{L}_r (t(x), s(x))
\end{equation}
where $t$ represents the teacher and $s$ represents the student, and $x$ is a sample ($x\sim x$) from the dataset.

While it is a possible approach to do a knowledge distillation on normalizing flows, it has a number of serious drawbacks.
First of all, distilling in this manner we would be limited to our dataset, meaning that we can only pass information to the smaller model on already known locations.
Another issue with this approach is that it is not making use of flow's invertible property, effectively treating a very complex model as a simple regression.
While this is useful, we are not transferring as much knowledge  as is possible from the teacher to the student. 

\subsection{Intermediate Latent Knowledge Distillation}

An extension to latent knowledge distillation is to use latent information from intermediate features of the flow.
Similar to the perceptual knowledge distillation~\cite{young2021featurealign}, one could link intermediate layers between the teacher and student flow blocks. 
% , analogous to figure \ref{fig:intermediate}. 
While this approach works just like the previous one, it has a nice benefit of constraining not only output but a student flow as a whole, making knowledge distillation ``stronger.'' 
Through this method we are able to pass significantly more information from the teacher network to the student.
We refer to this type of knowledge transfer as \textit{Intermediate Latent Knowledge Distillation} loss, or ILKD.

For an arbitrary flow step we can compare the latent information between the student and teacher, similar to equation~\ref{eq:LKD}. 
The difference is that this time we are using an arbitrary flow step.
We will refer to the teacher step as $t_i$ and the student step as $s_i$.
Similarly the metric may be arbitrary, but we use $L_1$.
%for each $j$-th intermediate layer's input $x_{j-1}$, for layers $\{l_1, l_2, \dots, l_m\}$ of student flow $s_{l_i}$ and their tied matches from teacher flow $t_{k l_i}$, where $k$ is a positive multiplier, \textbf{ILKD} could be defined as:
\begin{equation}
\label{eq:ILKD}
    \mathcal{L}_\text{ILKD}(t, s, x) = \sum_{i} \mathcal{L}_r (t_{i}(x), s_i(x))
\end{equation}

While many networks have explored the usage of $L_p$ losses in their networks,
Compositional Normalizing Flows have a unique property that subtly differentiate
this formulation.
At each layer the network must describe a probability distribution and since
each transformation is a change of variables, this results in transformations
that do not contain knots in their
trajectories~\cite{NEURIPS2019_5d0d5594,grathwohl2018ffjord,chen2018neural}.
$L_p$ losses calculate the differential between individual elements in the
latent representation, they place pressure on the networks to have the same
geometric representations within the latent space.
Due to the aforementioned properties of compositional Normalizing Flows, when
the $L_p$ loss is minimized, so will distributional losses like KL-Divergence.
We use $L_1$ loss due to the high dimensional nature of our data and the purpose
of ILKD is to ensure that the trajectories of the teacher and student models
align.
The goal is to compress $N$ flow steps into $M$, where $M<N$.

While we could use any arbitrary flow step to distill knowledge to, there are more optimal steps to transfer to. 
We choose to use flow steps that logically correspond to the progressive change of variables.
One simple example of this may be to have a student model have half the depth of the teacher and then every student flow step learns from every other teacher flow step.
This example would be akin to trying to having two flow steps in a teacher model distilled into a single flow step in the student model, thus halving the number of flow steps.
This ensures alignment within the trajectories and reduces potential
complications due to the biases of different flow step formulations.
Within our work we choose to distill this knowledge at each split level, finding that this leads to a balance of computation and complexity that each flow can learn.
%we believe that this is a good choice as the latent representations of similar performing networks should be similar.
Additionally, this allows for more flexibility, as it is reasonable that the flow steps will need different operations to reach the same latent representation in a different number of steps. 
Distilling at too frequent of intervals may over constrain the student model, making it inflexible to more complex distributions. 

%while both of these methods have advantages, both of the proposed frameworks are also limited by the size of the dataset. 
%ideally, we would like to have a method that is not limited by that and provides a way to distill knowledge on an entire learned data manifold rather than on a number of already seen samples.

\subsection{Synthesized Knowledge Distillation}

Another method to transfer knowledge is to look at the synthesized information from each network.
Not only do we want the latent information from the flows to be similar, but we also want the generated samples to be similar. 
To accomplish this simply by using random samples from the learned distribution and then comparing their synthesized results.
That is
\begin{equation}
\label{eq:SKD}
    \mathcal{L}_\text{SKD}(t, s, z) = \mathcal{L}_r(t^{-1}(z), s^{-1}(z))
\end{equation}
here we let $z$ represent the generated sample.
Additionally, we use $t^{-1}$ and $s^{-1}$ to represent the inverse flows, which are the generators as described in section~\ref{sec:nf}.
Unlike a cycle loss~\cite{cyclegan2017} we do not need to know the original image that was generated, just that we constrain the student's generation (backwards inference) path to be similar to that of the teacher's.
Additionally, unlike Baranchuk~\etal s distillation technique, we do not require that the generator be conditional.

\begin{table*}[!htb]
   \centering
    \caption{Averaged test log-likelihood (in nats) for unconditional density estimation (higher is better) across multiple runs. }

   %\resizebox{1.0\textwidth}{!}{
   \begin{tabular}{|l|l|c|c|c|c|c|}
      %\toprule
      \hline%\noalign{\smallskip}
     Architecture
     & Model
     & POWER
     & GAS
     & HEPMASS
     & MINIBOONE
     & BSDS300
     \\

     %\midrule
     \hline
    
    %  \midrule[0em]
    
     \multirow{4}{*}{GLOW} & Student
     & $-0.228$    & $5.967$    & $-22.668$  & $-17.251$  & $147.298$  \\
     & LKD Student
     & $-0.132$    & $6.008$    & $-22.332$  & $-17.136$  & $162.103$  \\
     & ILKD Student
     & $-0.133$    & $6.191$    & $-22.187$  & $-17.008$  & $163.148$  \\
     & SKD Student
     & $\mathbf{-0.078}$    & $\mathbf{6.515}$ & $\mathbf{-21.852}$  & $\mathbf{-16.130}$  & $\mathbf{163.953}$  \\
    
     %\midrule
     \hline
     GLOW & Teacher
     & \phantom{-}$0.143$    & $6.604$    & $-19.938$  & $-13.597$  & $165.702$  \\

    %  \midrule[0em]

    %  \bottomrule
    %  \midrule[0em]
    %\midrule
    \hline
    \hline
    
     \multirow{4}{*}{MAF} & Student
     & $-0.152$    & $4.385$    & $-21.904$  & $-15.314$  & $155.463$  \\
     & LKD Student
     & $-0.149$    & $4.473$    & $-21.389$  & $-15.217$  & $155.629$  \\
     & ILKD Student
     & $\mathbf{-0.145}$    & $\mathbf{4.502}$    & $\mathbf{-21.223}$  & $\mathbf{-15.184}$  & $\mathbf{155.785}$  \\
     & SKD Student
     & -    & -    & -  & -  & -  \\
    
     %\midrule
     \hline
     MAF & Teacher
     & $0.133$    & $5.887$    & $-20.662$  & $-13.488$  & $159.442$  \\

    %  \midrule[0em]

     %\bottomrule
     \hline
   \end{tabular}
   
   %\caption{Averaged test log likelihood (in nats) for unconditional density estimation (higher is better) across multiple runs. $d$ is the number of dimensions and $N$ is the size of the dataset. Student models have the same architecture as the distilled models, but were trained without any knowledge distillation losses.}
%   \vspace{1em}
   \label{table:tabular-metrics}

\end{table*}

\begin{table*}[!htb]
   \centering
   \caption{Time consumption for a single batch inference averaged across multiple batches and the number of parameters (in thousands). Average time (ms) and number of parameters (in thousands) are reported.}
   %\resizebox{1.0\textwidth}{!}{
   \begin{tabular}{|l|l|c|c|c|c|c|c|}
     %\toprule
     \hline
     Arch%itecture
     & Model
     & Metric
     & POWER
     & GAS
     & HEPMASS
     & MINIBOONE
     & BSDS300
     \\
    
     %\midrule
     \hline
    
    %  \midrule[0em]
    
     \multirow{4}{*}{GLOW} & \multirow{2}{*}{Student} & Time (ms)
     & $2.32 \pm 0.16$    &  $2.46 \pm 0.1$   & $2.55 \pm 0.35$  & $2.47 \pm 0.07$  & $2.45 \pm 0.07$  \\
     %\cline{3-8}
     &  & Params (K)
     & $13.8$    &  $14.2$   & $17.4$  & $24.9$  & $34.4$  \\
     \cline{2-8}
     & \multirow{2}{*}{Teacher} & Time (ms)
     & $3.65 \pm 0.26$    & $3.88 \pm 0.09$    & $4.41 \pm 0.28$  & $3.95 \pm 0.11$  & $3.89 \pm 0.14$  \\
     %\cline{3-8}
     &  & Params (K)
     & $86.7$    & $87.8$    & $96.3$  & $114.2$  & $134.7$  \\

    % \midrule
    \hline
    %  \midrule[0em]
    %  \bottomrule
    %  \midrule[0em]
    
     \multirow{4}{*}{MAF} & \multirow{2}{*}{Student} & Time (ms)
     & $2.0 \pm 0.21$    & $1.98 \pm 0.19$    & $1.82 \pm 0.05$  & $1.82 \pm 0.05$  & $1.91 \pm 0.22$  \\
     &  & Params (K)
     & $5.0$    & $5.6$    & $9.4$  & $15.9$  & $21.8$  \\
    \cline{2-8}
     & \multirow{2}{*}{Teacher} & Time (ms)
     & $3.34 \pm 0.22$    & $3.22 \pm 0.18$    & $3.34 \pm 0.23$  & $3.36 \pm 0.26$  & $3.45 \pm 0.2$  \\
     &  & Params (K)
     & $10.1$    & $11.2$    & $18.9$  & $31.8$  & $43.6$  \\

    %  \midrule[0em]

     %\bottomrule
     \hline
   \end{tabular}
   %}
   
%   \vspace{1em}
   \label{table:tabular-perfomance}

\end{table*}

Since we cannot count on the whole latent space to be as equally covered by the flow mapping, the stability of this method requires that the latent spaces of the student and teacher are similar to each other. 
In other words, we should distill the teacher's latent space into the student's one before we start the procedure.
We note that we found that this method is often unstable.
We believe that this is due to NFs difficulty in backwards inference, notably in that they tend to have poorer sampling quality.
We believe that this is still a useful notion and will become more stable as the quality of flow's image generation increases.

\subsection{Flow Distillation}
we can combine all these types of distillations together and create a much stronger from of distillation that each provides independently.
The hyperparameters $\lambda_i$, representing the weight of the distillation, the resulting loss can be written as follows:
\begin{equation}
\label{eq:SKD_total}
\begin{split}
    \mathcal{L}(t, s, x, z) &= \lambda_0\log{p_s(x)}\\ 
    &+ \lambda_1 \mathcal{L}_\text{(I)LKD}(t, s, x)\\
    &+ \lambda_2 \mathcal{L}_\text{SKD}(t, s, z)
    \end{split}
\end{equation}
here we use $l_{(i)LKD}$ to denote both the intermediate and standard latent knowledge distillation, noting that the final latent step is just another ``intermediate'' step.
We show an example of the full distillation in figure~\ref{fig:distill}.

%\vspace{-1cm}
\section{Experiments}
\label{sec:experiments}

%The experiments were conducted on an internal GPU cluster. 
All calculations were performed on a single GPU Tesla V100.

\subsection{Models}

%To demonstrate that the proposed method's performance is model agnostic, several flow-based models were selected.
%For the experiments we used Masked Autoregressive Flow~\cite{papamakarios2017masked} and GLOW ~\cite{kingma2018glow} as previously discussed in Section ~\ref{sec:related}.
To demonstrate that this method is model agnostic we demonstrate by using Masked Autoregressive Flow~\cite{papamakarios2017masked} and GLOW~\cite{kingma2018glow}, as previously discussed in Section~\ref{sec:related}.
In all experiments the teacher and students share the same basic model but differ in the number of flow steps, with the student model being smaller than the teacher.
For example, in the GLOW model both student and teacher share the number of levels (splits) but have a differing number of flow steps between them.
These models have differing architectures but also form the basis of many other types of flow models and thus we believe stand as good proxies.

\subsection{Tabular data}

We perform density estimation experiments on five standard tabular datasets that include four datasets from the UCI machine learning repository ~\cite{UCI} and on a dataset of natural image patches BSDS300 ~\cite{MartinFTM01}.
In Table~\ref{table:tabular-metrics}, we report the average log-likelihood on held-out test sets. 
For each model, we also provide the statistics for time and memory consumption in table~\ref{table:tabular-perfomance}.

%\vspace{-0.5cm}
%\vspace{-1cm}
\begin{figure*}[ht]
    \begin{subfigure}[b]{0.30\linewidth}
        \includegraphics[width=\textwidth]{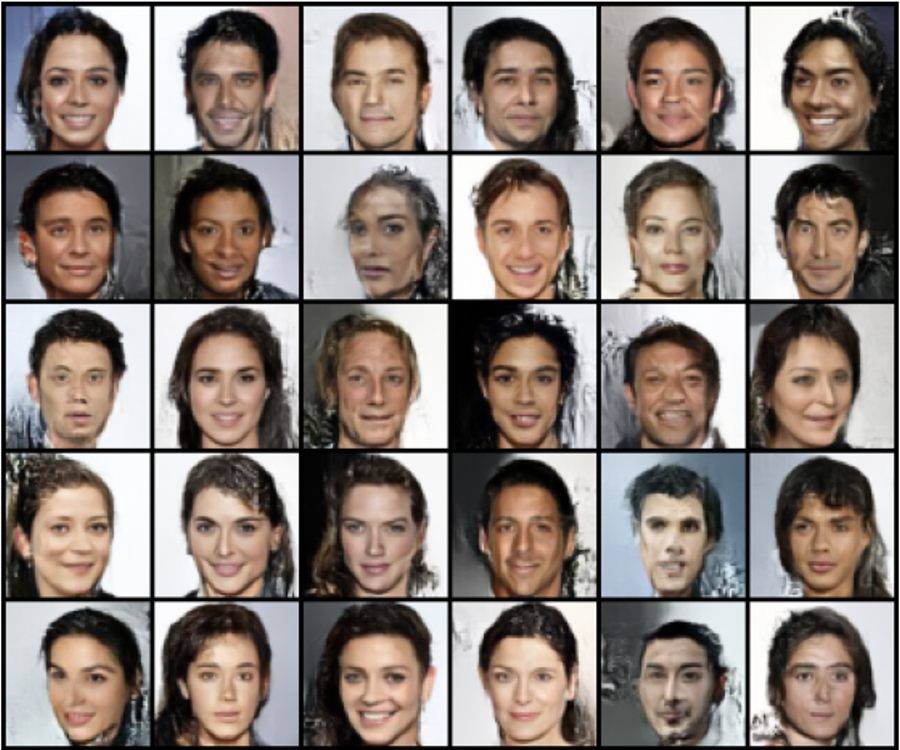}
        \caption{Teacher}
        \label{fig:teacher}
    \end{subfigure}
    \hfill
    \begin{subfigure}[b]{0.30\linewidth}
        \includegraphics[width=\textwidth]{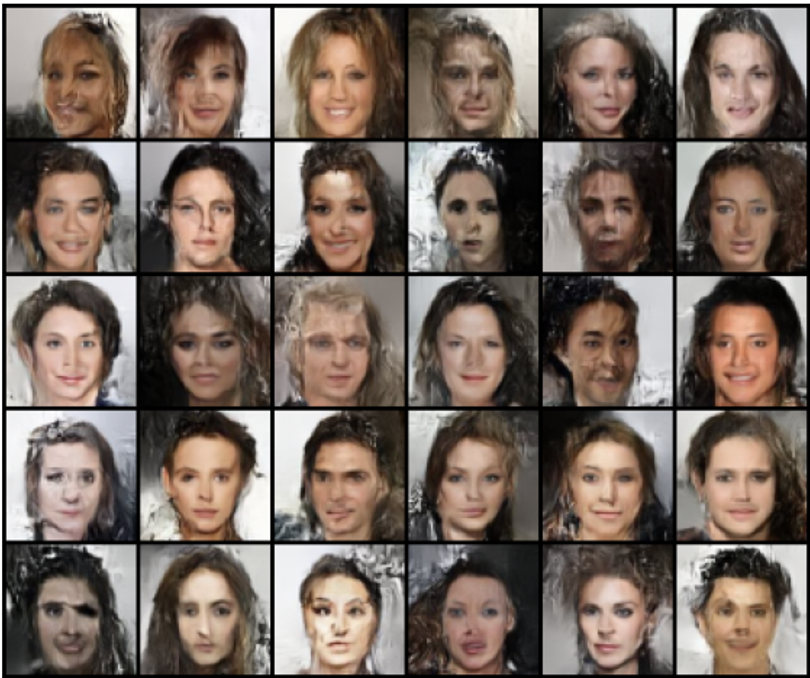}
        \caption{With KD}
        \label{fig:withkd}
    \end{subfigure}
    \hfill
    \begin{subfigure}[b]{0.30\linewidth}
        \includegraphics[width=\textwidth]{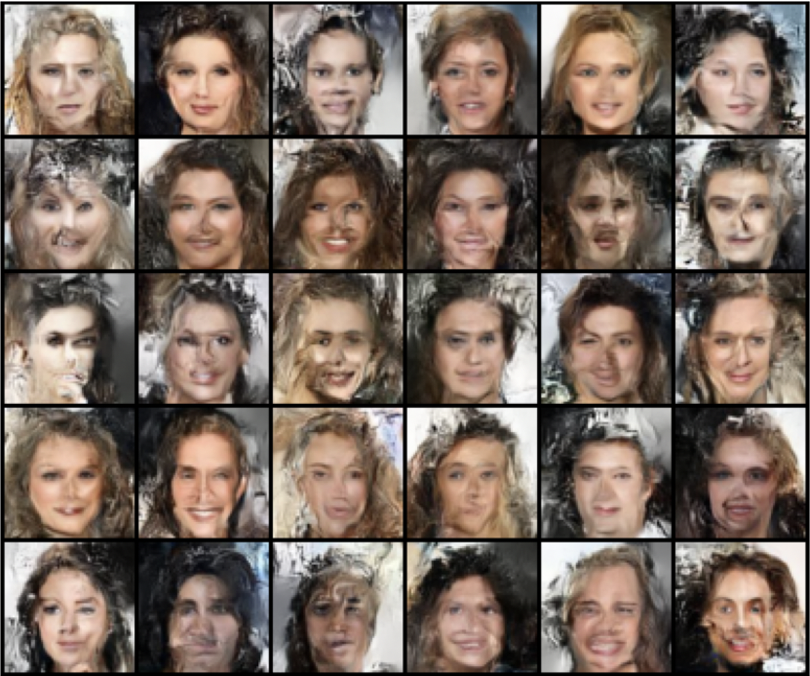}
        \caption{Without KD}
        \label{fig:nokd}
    \end{subfigure}
    \caption{CelebA samples from teacher model (\ref{fig:teacher}), student model (\ref{fig:withkd}), and student model with no knowledge distillation (\ref{fig:nokd}). All images are generated at 64$\times$64 resolution and with temperature=$0.7$.}
    \label{fig:celeba}
\end{figure*}

\begin{figure*}[ht]
    \centering
    \begin{subfigure}[b]{0.30\linewidth}
        \includegraphics[width=\textwidth]{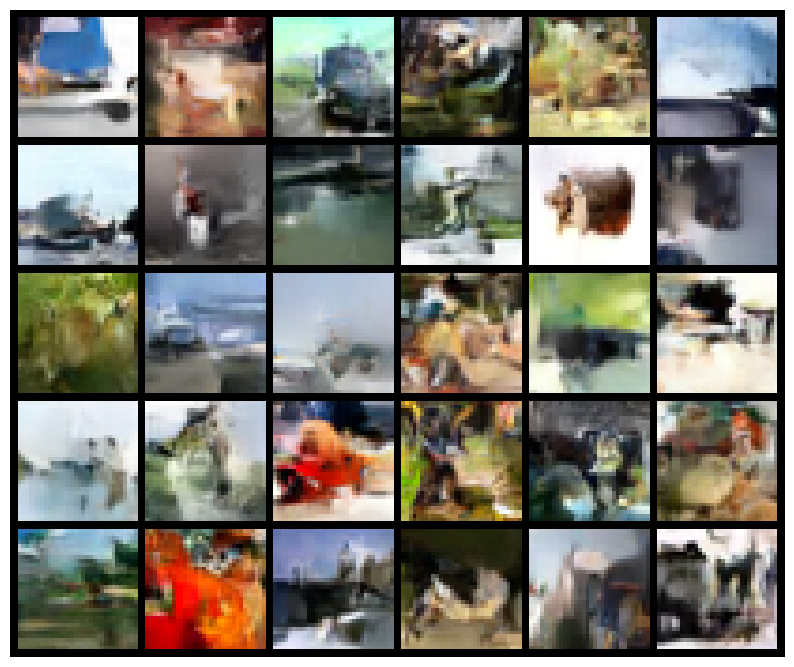}
        \caption{Teacher}
        \label{fig:cifarTeacher}
    \end{subfigure}
    \hfill
    \begin{subfigure}[b]{0.30\linewidth}
        \includegraphics[width=\textwidth]{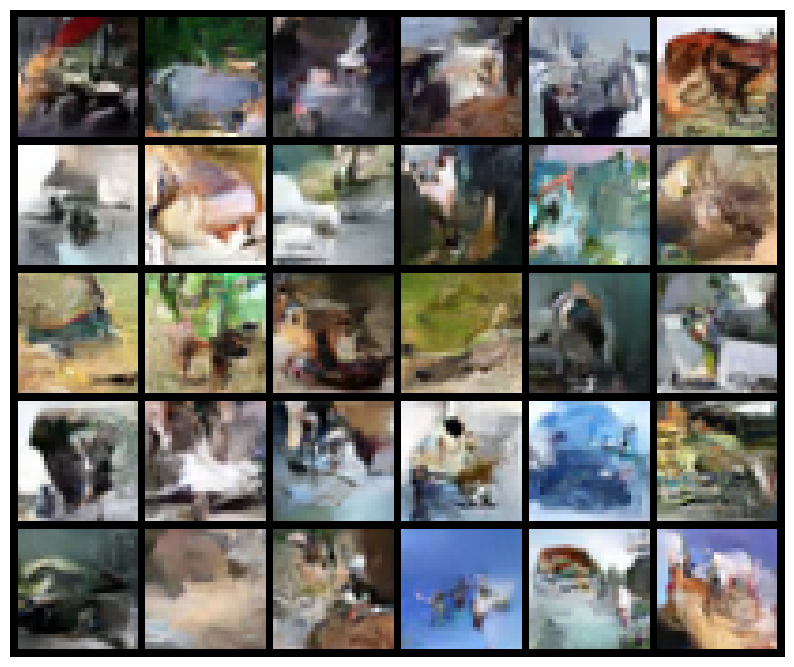}
        \caption{With KD}
        \label{fig:cifarwKD}
    \end{subfigure}
    \hfill
    \begin{subfigure}[b]{0.30\linewidth}
        \includegraphics[width=\textwidth]{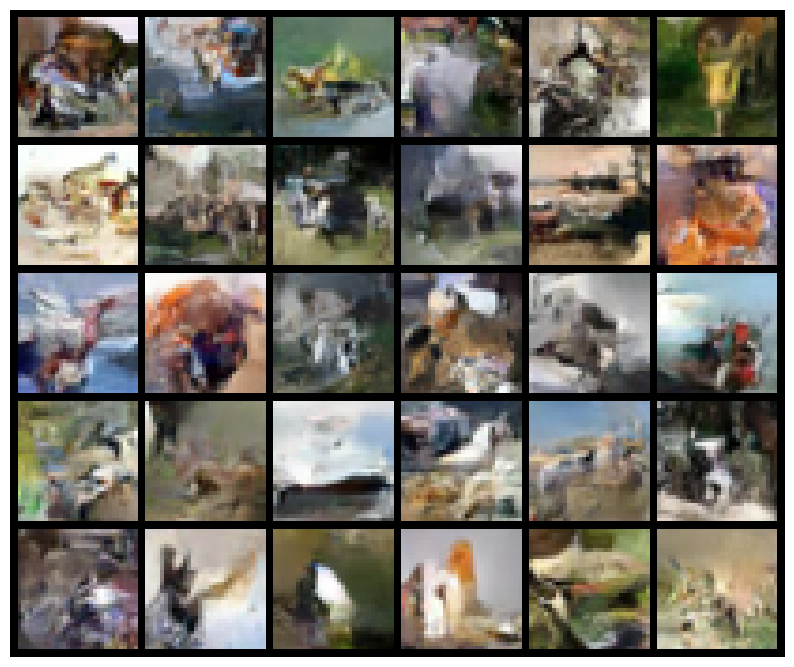}
        \caption{Without KD}
        \label{fig:cifarwoKD}
    \end{subfigure}
    \caption{CIFAR-10 samples from teacher model (\ref{fig:cifarTeacher}), student model (\ref{fig:cifarwKD}), and student model with no knowledge distillation (\ref{fig:cifarwoKD}). All images are generated at $32\times32$ resolution and with a temperature of 0.7.}
    \label{fig:cifar}
\end{figure*}

As can be seen in table~\ref{table:tabular-perfomance}, the proposed LKD and the improved ILKD methods both allows for a significant reduction in time and memory consumption.
The SKD method not only improves student flow performance on all datasets but also has better quality than other distillation methods for the GLOW teacher and student. 
However, its numerical instability leads to a discrepancy in the MAF student's training with SKD.
We believe that this is due to MAF's weaker modeling power, noting that this model is substantially smaller and significantly under performs compared to the GLOW based models.

We train each flow for $1\times10^4$ iterations with a batch size of 65,536
elements and a learning rate of $5\times10^{-5}$ on tabular data and AdamW~\cite{loshchilov2019decoupled} as an optimizer. 
In Table~\ref{table:tabular-perfomance} the rows labeled with LKD we use $\lambda_0=1$ ($\lambda_1=\lambda_2=0$), for ILKD we use $\lambda_0=0.9$ and $\lambda_1=0.1$ ($\lambda_2=0$), and for our SKD we use $\lambda_0=0.85$ and $\lambda_1=\lambda_2=0.075$, as described in Equation~\ref{eq:SKD_total}. 
We found that these combinations of weights are stable and improve the quality of distilled models for all datasets.
We also note that as with most NFs, training and stability are significantly affected by batch sizes.
Additionally, Flow based networks tend to learn best when gradients are not rapidly changed, often requiring longer warmups and gradient clipping.
We note here that this is likely a reason that the optimal distillation weights are small for ILKD and SKD experiments.
Small perturbations in flow training can often lead to compounding downstream changes.
A detailed overview of network parameters can be found in table~\ref{table:tabular-config}.

\begin{table}[!htb]
  \centering
  \caption{Model configurations for generation of tabular data. Provided for GLOW and MAF architectures. Number of levels (L) is equal to 1. Notation is taken from the original paper \cite{kingma2018glow}. }
  \begin{tabular}{|l|c|c|}
    %\toprule
    \hline
    \textbf{GLOW}% & \multicolumn{2}{c}{}
    %& \multicolumn{2}{c}{MAF}
    %\\
    & Level (L)
    & Hidden
    %& Depth (K)
    %& Hidden
    \\
    % \midrule[0em]
    %\midrule
    \hline
    
    % \midrule[0em]
    
    Student
    & $3$    & $32$\\%  & $3$  & $32$  \\
    Teacher
    & $3$    & $64$\\% & $6$  & $32$  \\

    %\midrule
    \hline
    \textbf{MAF} & Depth (K) & Hidden\\% \multicolumn{2}{c}{}\\
    \hline
    Student & $3$ & $32$\\
    Teacher & $6$ & $32$\\

    % \midrule[0em]

    %\bottomrule
    \hline
  \end{tabular}
%   \vspace{1em}
  \label{table:tabular-config}

\end{table}

%\vspace{-0.5cm}

% add more flex
%\vspace{-1cm}
\subsection{Image data}

To demonstrate our method's performance for convolution-based normalizing flows, we used CelebA~\cite{liu2015faceattributes} and CIFAR-10~\cite{Krizhevsky09learningmultiple} image datasets.
In the following experiments, we used GLOW~\cite{kingma2018glow} for the unconditional image generation.  
Both of these experiments use the affine coupling layer as discussed in the GLOW paper.
Here we use the same training parameters as the tabular data but with a batch size of 32.

We found that the SKD increased the quality of the student models, but also found that this method was unstable. We, again, believe that this is due to the low sampling quality of these networks.
Because of this we focus on just using ILKD, as this is both stable and consistently increases the quality of generated samples. 

Configurations of trained models are provided in table~\ref{table:image-config}. 
Additionally, 
    Figures~\ref{fig:celeba} 
    and 
    ~\ref{fig:cifar} 
show random samples from the teacher models as well as students with and without knowledge distillation.
%the student trained without KD, the student trained with ILKD and the teacher.
We can see that the students with knowledge distillation significantly outperform students without knowledge distillation.
This reflects the results we saw in the table~\ref{table:image-metrics}. 
Here we note, as seen in Table~\ref{table:image-config} that the CIFAR student is a quarter the size of the teacher yet has a $<2\%$ difference in sampling quality, an improvement of $3\%$ over the non-distilled student's FID.
Similarly, the CelebA's student is approximately one eighth of the teacher model and over $35\%$ improvement over the baseline student.
This directly demonstrates that these distillation techniques have a powerful effect on student models.

\begin{table}[!htb]
  \centering
  \caption{Metrics for the image generation task for the GLOW architecture using ILKD on the test set: bits per dimension and FID (lower is better).}

  \begin{tabular}{|l|c|c|c|c|}
    %\toprule
    \hline
    & \multicolumn{2}{c|}{CIFAR-10}
    & \multicolumn{2}{c|}{CelebA}
    \\
    \cline{2-5}

    & bpd
    & \multicolumn{1}{c|}{FID}
    & bpd
    & FID
    \\
    % \midrule[0em]
    %\midrule
    \hline
    
    % \midrule[0em]
    
    Student &
    $3.498$ & \multicolumn{1}{c|}{$71.177$}  & $2.479$  & $68.127$  \\
    ILKD Student &
    \textbf{3.481}  & \multicolumn{1}{c|}{\textbf{69.371}}  & \textbf{2.475}  & \textbf{54.480}  \\
    % \midrule[0em]
    %\midrule
    \hline
    % \midrule[0em]
    Teacher &
    $3.423$  & \multicolumn{1}{c|}{$68.503$}  & $2.474$  & $37.460$  \\
    % \midrule[0em]

    %\bottomrule
    \hline
  \end{tabular}
%   \vspace{1em}
  \label{table:image-metrics}
\end{table}

\begin{table}[!htb]
  \centering
  \caption{Model configurations for CIFAR-10 image generation tasks (GLOW). Notation is taken from the original paper \cite{kingma2018glow}.}
%   \vspace{1em}
  \label{table:image-config}
  \begin{tabular}{|l|c|c|c|c|}
    \hline
    %\midrule
    & Levels (L)
    & Depth (K)
    & Hidden
    & Params\\
    \hline
    \textbf{CIFAR-10} & \multicolumn{4}{c|}{}\\
    \hline
    % \midrule[0em]
    
    % \midrule[0em]
    
     Student
     & $8$    & $3$    & $512$  & \multicolumn{1}{c|}{$\phantom{0}11$M}  \\
     Teacher
     & $32$    & $3$    & $512$  & \multicolumn{1}{c|}{$44.2$M} \\
     \hline
     \textbf{CelebA} & \multicolumn{4}{c|}{}\\
     \hline
     %\midrule[0em]
     %\midrule
    
     %\midrule[0em]
    
     Student
     &$16$    & $3$   & $256$  & $\phantom{0}7.9$M  \\
     Teacher
     & $32$   & $3$     & $512$  & $61.2$M  \\

    %  \midrule[0em]

     \hline
  \end{tabular}

\end{table}

\subsection{Latent Space Corruption}
To ensure the knowledge distillation does not corrupt the hidden space, we need to ensure that random samples from the students still maintain similar quality images.
With high dimensional information, it is possible for Normalizing Flows, and other models, to have a small KL-Divergence but also have poor sampling quality.
However, high quality samples can only happen if there is a sufficiently good enough cover within the learned latent space.
Thus, we propose to measure the quality of the inferred samples for randomly chosen images $\boldsymbol{u}, \boldsymbol{v}$ and an $\alpha \in [0,1]$.
The preserved norm of the latent vector can be defined as:
%that the hidden space is not corrupted because of the knowledge distillation, 
%we propose conducting the following experiment. As the high-quality interpolation between latent vectors of real objects is possible only when Normalizing Flow is trained well and the mappings sufficiently tightly cover latent space, we propose to measure the quality of random interpolations with preserved norm~\ref{eq:norm_interpolation}. For randomly chosen images $\boldsymbol{u, v}$ and $\alpha \in (0, 1)$, their interpolated latent vector with preserved norm is defined as:

\begin{equation}
\begin{split}
    \boldsymbol{f(u, v, \alpha)} &= ((1 - \alpha) \boldsymbol{f(u)} + \alpha \boldsymbol{f(v)}) \\
    &\cdot \frac{(1 - \alpha) ||\boldsymbol{f(u)}|| + \alpha ||\boldsymbol{f(v)}||}{|| (1 - \alpha) \boldsymbol{f(u)} + \alpha \boldsymbol{f(v)} ||}
    \end{split}
\label{eq:norm_interpolation}
\end{equation}

The results of this method are provided for CelebA dataset in Table~\ref{table:interpolation-metrics}.
In this table we can see that the ILKD Student performs significantly better than the student without knowledge distillation.
This is especially true for a temperature of $0.7$, which generates better samples on all models.
A similar temperature was found to have better performance in the original GLOW paper.
In Figures~\ref{fig:celeba} and \ref{fig:cifar}
we can also see that the distilled models produce significantly higher quality samples than the non-distilled student models.
We show the resultant FID scores for these different temperatures in Table~\ref{table:interpolation-metrics}.
We note here the significant improvements 
%Here should be a table 4 with interpolation metrics, someone please fix it

 \begin{table}[!htb]
 \centering
    \caption{CelebA FID values of images obtained by interpolation in the latent space of trained models.}

   \centering
   \begin{tabular}{|l|c|c|}
      \hline
     & \multicolumn{2}{c|}{FID} \\
     \cline{2-3}
     & temp 1.0
     & temp 0.7
     \\
    %  \midrule[0em]
     \hline
    
    %  \midrule[0em]
    
     Student &
     $40.159$  & $28.432$  \\
     ILKD Student &
     \textbf{28.413}  & \textbf{19.688}  \\
    %  \midrule[0em]
     \hline
    %  \midrule[0em]
     Teacher &
     $19.062$  & $16.382$  \\
    %  \midrule[0em]

     \hline
   \end{tabular}
   
%   \vspace{1em}
   \label{table:interpolation-metrics}
\end{table}

%\vspace{-1cm}

\section{Conclusion}
\label{sec:conclusion}

In this work, we demonstrated a novel methods for normalizing flow knowledge
distillation, taking advantage of their unique properties. We demonstrated that
on a variety of different types of datasets that the methods significantly
improve the performance of every flow, greatly decreasing the sampling quality
gap between teacher and student flows.% by $\approx80\%$.

Compared to other distillation methods, ours utilize the unique properties of
the normalizing flows' invertibility for better quality and performance. This
allows for high quality students to be trained in a simple and efficient manner
with minimal loss to sampling performance. Additionally, the resulting
probability properties of a distilled flow are kept, making our method a
straightforward application to normalizing flow distillation.

%We invite other authors to test proposed improvements for the method and extend its application to other domains.

% add flex about the fact that we are first to do any distillation except waveтet

{
    \small
    \bibliographystyle{ieeenat_fullname}
    \bibliography{main}
}

\end{document}